\documentclass[conference]{IEEEtran}
\usepackage{cite}
\usepackage{amsmath,amssymb,amsfonts}
\usepackage{algorithmic}
\usepackage{graphicx}
\usepackage{textcomp}
\usepackage{xcolor}
\usepackage[caption=false, font=footnotesize]{subfig}
\usepackage{multirow}
\usepackage{colortbl}
\usepackage{xcolor}
\usepackage{listings}
\usepackage{fancybox} 
\usepackage{fancyvrb}
\usepackage{hyperref}
\usepackage{booktabs}
\usepackage{adjustbox}

\renewcommand{\textit}[1]{\emph{#1}}

\begin{document}

\title{Missed Connections: Lateral Thinking Puzzles for Large Language Models}

\author{
    \IEEEauthorblockN{Graham Todd}
    \IEEEauthorblockA{\textit{New York University Tandon}\\
    Brooklyn, New York, USA \\
    gdrtodd@nyu.edu}
    
    \and
    
    \IEEEauthorblockN{Tim Merino}
    \IEEEauthorblockA{\textit{New York University Tandon}\\
    Brooklyn, New York, USA \\
    tm3477@nyu.edu}
    
    \and
    
    \IEEEauthorblockN{Sam Earle}
    \IEEEauthorblockA{\textit{New York University Tandon}\\
    Brooklyn, New York, USA \\
    se2161@nyu.edu}
    
    \and
    
    \IEEEauthorblockN{Julian Togelius}
    \IEEEauthorblockA{\textit{New York University Tandon}\\
    Brooklyn, New York, USA \\
    julian@togelius.com}
}

\maketitle

\begin{abstract}
The \textit{Connections} puzzle published each day by the New York Times tasks players with dividing a bank of sixteen words into four groups of four words that each relate to a common theme. Solving the puzzle requires both common linguistic knowledge (i.e. definitions and typical usage) as well as, in many cases, lateral or abstract thinking. This is because the four categories ascend in complexity, with the most challenging category often requiring thinking about words in uncommon ways or as parts of larger phrases. We investigate the capacity for automated AI systems to play \textit{Connections} and explore the game's potential as an automated benchmark for abstract reasoning and a way to measure the semantic information encoded by data-driven linguistic systems. In particular, we study both a sentence-embedding baseline and modern large language models (LLMs). We report their accuracy on the task, measure the impacts of chain-of-thought prompting, and discuss their failure modes. Overall, we find that the \textit{Connections} task is challenging yet feasible, and a strong test-bed for future work. 
\end{abstract}

\begin{IEEEkeywords}
Language models, reasoning, AI, evaluation
\end{IEEEkeywords}

\section{Introduction}

\textit{Connections} is a daily word puzzle game published by the New York Times which debuted on June 12th, 2023\footnote{See and play examples here: \href{https://www.nytimes.com/games/connections}{https://www.nytimes.com/games/connections}}. The game consists of a grid of sixteen words that must be separated into four groups of four related words. These groups are determined by the puzzle, and each group is unified by a particular theme or clue. The groups in each puzzle are also sorted into four levels of increasing complexity. For instance, a ``simple'' group might be \textit{FISH} and consist of the words \textit{bass}, \textit{flounder}, \textit{salmon}, and \textit{trout}. By contrast, a ``tricky'' group might be \textit{FIRE \_\_\_} (i.e. ``words that complete a phrase starting with the word \textit{fire}) and consist of the words \textit{ant}, \textit{drill}, \textit{island}, and \textit{opal}. 

In order to solve a \textit{Connections} puzzle, players must both understand the literal meanings of the sixteen words (e.g. that \textit{bass} can refer to a type of fish, among its definitions) in addition to some of the subtleties of their usage in context (e.g. that there is a variety of \textit{opal} called a \textit{fire opal}). Some particularly challenging groupings might even refer to non-semantic properties of words: one recent puzzle relies on noticing that words like \textit{MOW} and \textit{NOON} read the same when rotated by 180 degrees. Of course, a given word affords countless semantic and non-semantic groupings---the real challenge of a \textit{Connections} puzzle is finding shared properties among words and evaluating their likelihood in the context of potential distractors (i.e. words that appear to belong to one category but are actually assigned to another). In this way, the puzzle acts as both a test of linguistic understanding and abstract reasoning.

\begin{figure}
    \centering
    \includegraphics[width=0.65\columnwidth]{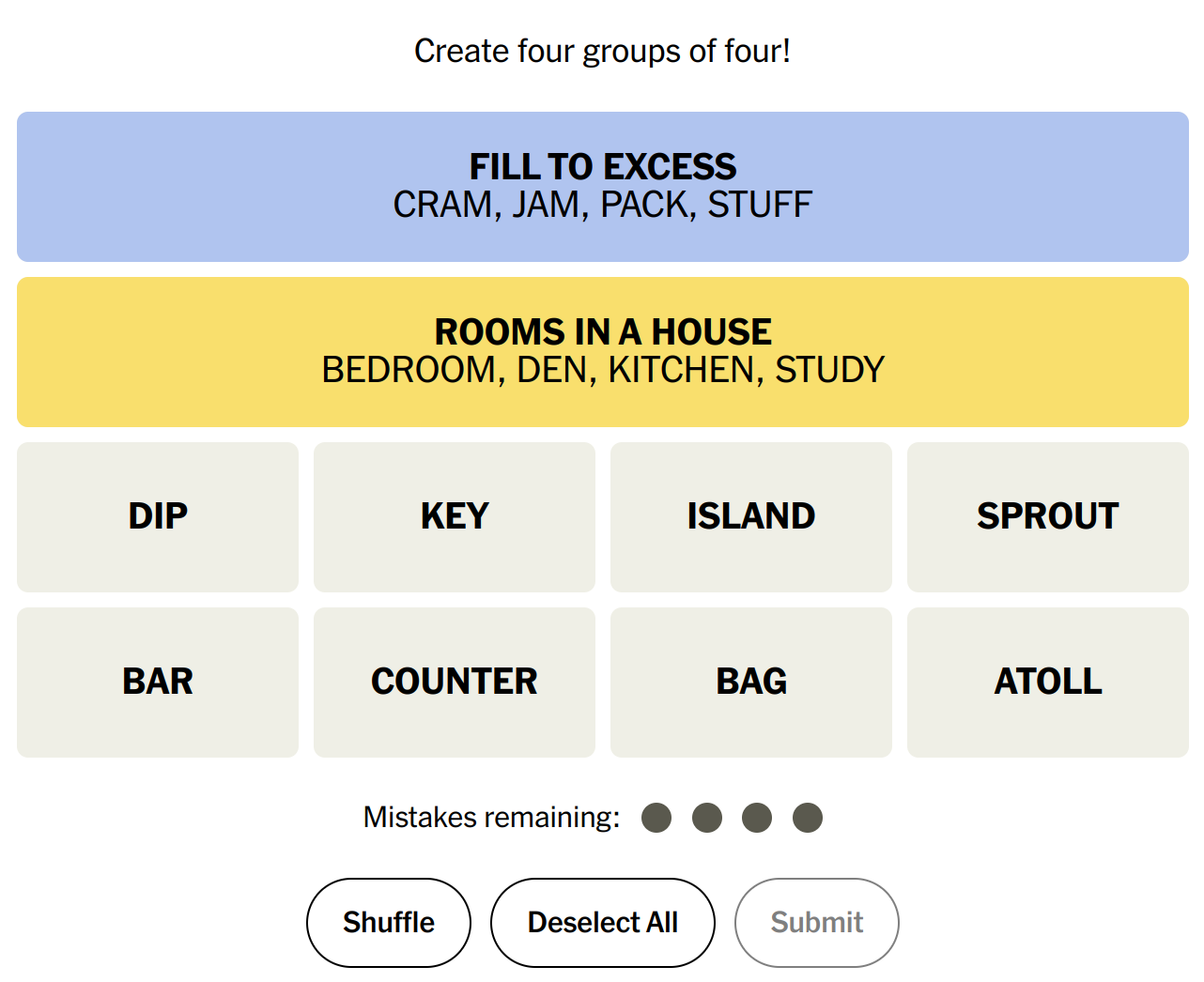}
    \caption{An example \textit{Connections} puzzle taken from the New York Times web interface on November 28th, 2023}
    \label{fig:connections_puzzle}
\end{figure}

Large language models (LLMs) and other data-driven approaches in natural language processing (NLP) attempt to capture this kind of linguistic information by mining patterns in vast corpora of human-produced text. This approach has proven very successful in a variety of domains~\cite{reimers-2019-sentence-bert,brown2020language,thirunavukarasu2023large, liu2023agentbench}, and recent work has begun to investigate the capacity of LLMs in particular to perform multi-step abstract reasoning~\cite{wei2022chain,lu2022learn,suzgun2022challenging, webb2023emergent}. We propose the \textit{Connections} puzzle as a way to study these capabilities in modern systems, among the array of existing tasks and benchmarks. We investigate whether the latent information encoded by these data-driven approaches is sufficient to solve the puzzles by using clustering-based algorithms on sentence embeddings and by directly querying the Generative Pretrained Transformer (GPT) family of large language models. We find that these methods are able to solve the puzzles a sizeable proportion of the time, but with far from perfect accuracy. This performance gap provides a fertile ground for studying both the successes and failures of modern systems to encode and retrieve semantic information. We also examine the impacts of different prompts on the performance of the LLMs (finding a notable effect) and examine a more challenging variant of the basic puzzle. We conclude by discussing a variety of promising avenues for future work.

\section{Related Work}

As LLMs have improved in performance and surged in popularity, a current of research has emerged that examines their capabilities to perform a variety of complex and abstract tasks viewed through the lens of linguistic sequence generation. In the domain of games, recent work has demonstrated that LLMs can be capable players~\cite{wang2023voyager,tsai2023can,meta2022human,noever2020chess,ciolino2020go}, non-player characters (NPCs)~\cite{urbanek2019learning,park2023generative} and content generators \cite{summerville2016super,sudhakaran2024mariogpt,todd2023level,wang2023bytesized32}. 

Our work follows prior efforts at using LLMs and other NLP systems specifically in the context of text-based puzzles. One approach develops and evaluate player agents for the language game \textit{Codenames} based on TF-IDF scores, the Naive-Bayes algorithm, and GPT-2 word embeddings \cite{Jaramillo_Charity_Canaan_Togelius_2020}. They find that only the agent built on the language model embeddings achieves competitive results. Another approach uses LLMs to solve and generate the word-based \textit{Sunday Puzzles} published each week by National Public Radio (NPR) \cite{zhao2023solving}. In a multiple-choice setting, they find that GPT-3.5 achieves up to 50\% accuracy but note that the model struggles significantly when tasked with generating interesting and novel puzzles. Compared to the \textit{Connections} puzzles, the weekly puzzles from NPR tend to rely more on character-level transformations of words and relatively obscure references, and less on common-sense associations. Finally, an existing work studies the use of ``cryptic crossword'' clues as an NLP benchmark \cite{rozner2021decrypting}. The authors use the T5 language model, which exhibits poor performance but is improved via the use of a specialized training curriculum. We continue this general effort by using the \textit{Connections} puzzle to study the information contained in latent representations of words used by modern LLMs.

Most directly, we build off of preliminary prior work that examines the ability of ChatGPT (an instance of GPT-3.5) to solve \textit{Connections} puzzles \cite{perowne2023connections}. The authors examine the first 60 puzzles and task the model with producing all four groups at once (though see discussion in \autoref{sec:challenge-results}). They note that the model tends to struggle with particular categories, such as those that tie words together which belong to a longer phrase (e.g. \textit{FIRE \_\_\_}). We systematize this approach by implementing an automated pipeline for solving and examining a larger set of \textit{Connections} puzzles, laying the groundwork for future research and demonstrating the puzzle's utility as an LLM benchmark. We also examine new models (a sentence embedding baseline and GPT-4) and provide an analysis of both the base puzzle and a more challenging variant. 

\section{Game Details}
\label{sec:details}
The \textit{Connections} puzzle is made up of a grid of sixteen English words (see \autoref{fig:connections_puzzle}). The initial arrangement of words in the grid is fixed, but it can be arbitrarily re-shuffled by the player. In order to submit a guess, the player selects a set of exactly four words. If the guess is correct (i.e. the four words belong to the same category) then the category is revealed and the words are removed from the grid. Each of the categories is assigned a color, and there is a stated ordering to category difficulty: yellow is the easiest, followed by green, blue, and purple. If three of the four words belong to the same category, then this fact is indicated to the player, but neither the words nor the category are revealed. If two or fewer words belong to the same category then no additional information is provided to the player. If the player makes four incorrect guesses then the puzzle is failed.

We implement the \textit{Connections} puzzle as described above with a \textsc{Python} interface, including the informational messages provided to the player but omitting the option to shuffle the order of the words. We also implement a slight variant of the \textit{Connections} puzzle. In this variant, all four groups must be submitted simultaneously and the only feedback is whether all four groups are correct. This is substantially more difficult version of the base game, since the ``tricky'' groups cannot be easily found by solving the simpler groups and thereby reducing the space of possible guesses.

\section{Data}
We collect a dataset of 250 puzzles from an online archive\footnote{Archive available here: \href{https://connections.swellgarfo.com/archive}{https://connections.swellgarfo.com/archive}}---representing daily puzzles from June 12th, 2023 to February 16th, 2024. Our methods do not use training data in order to improve performance---we use all 250 games for evaluation and do not split into validation or test sets. The utility of using some puzzles as training data remains an interesting question for future work.  

\section{Approaches}

\subsection{Sentence Embeddings}

Our first approach makes use of \textit{sentence embeddings}---high-dimensional vectors designed to encode semantic information contained in natural language. We make use of a variety of embedding models from the \textit{SentenceTransformers} library, specifically embeddings derived from \textsc{BERT} \cite{devlin2018bert}, \textsc{RoBERTa} \cite{liu2019roberta}, \textsc{MPNet} \cite{song2020mpnet}, and \textsc{MiniLM} \cite{wang2020minilm}.

In order to generate a guess, we first enumerate all $\binom{16}{4} = 1,820$ possible four-word groups. For each group, we compute the set of pairwise cosine similarities between the embeddings of each entry. We then sort the groups by their average pairwise cosine similarity in descending order and return the first group. If this guess matches one of the categories, then we repeat the process with the remaining 12 words (and beyond, if subsequent guesses are also correct). If the guess is incorrect, then we proceed to the group with the next-highest aggregate cosine similarity, continuing until either the puzzle is solved or we run out of guesses.

For the \textit{Connections} challenge variant described in \autoref{sec:details} in which all four guesses must be submitted simultaneously, we instead enumerate all $\frac{\binom{16}{4} \binom{12}{4} \binom{8}{4} \binom{4}{4}}{4!} = 2,627,625$ possible guesses consisting of four disjoint groups of four words. As above, we compute the within-group average pairwise cosine similarity and sort guesses by the sum of the aggregated similarities for each of their groupings. We input the highest-scoring guess and proceed down the list until either the puzzle is solved or we have run out of guesses. 

\subsection{Language Models}

Our second approach makes use of \textit{instruction-tuned large language models} in the GPT family released by OpenAI \cite{vaswani2017attention, openai2023gpt4}. We use the OpenAI API in order to query the large language models by providing the instructions for how to play \textit{Connections} as well as the current game state in natural language and requesting a guess in return. The game state information provided in the prompt includes the set of remaining words, any categories that have already been guessed, a list of previously guessed groupings, and the number of remaining guesses. After the initial guess, we also provide the model with the last response from the game (i.e. if a a guess was incorrect, or nearly correct) and its conversation history. Our complete set of prompts is available in the \hyperref[sec:prompts]{Appendix}. We parse out the model's response using a regular expression to obtain either a set of four words (in the standard game) or four sets of four words (in the variant game) and automatically input them into the puzzle as a guess, proceeding until the puzzle is solved or we run out of guesses as usual. We note that, unlike the sentence-embedding models, it's possible for the LLM-based approach to produce malformed guesses (e.g. words that do not appear in the puzzle). In these cases we provide a prompt to the model explaining that the guess was malformed and allow it to guess again. After 5 invalid guesses we terminate the puzzle and treat it as unsolved.

The specific versions of the LLMs we use are \textsc{gpt-3.5-turbo-1106} (referred to as \textsc{gpt-3.5-turbo}) and \textsc{gpt-4-1106-preview} (referred to as \textsc{gpt-4-turbo}). The more recently-trained of these two models, \textsc{gpt-4-turbo}, has a ``knowledge cut-off date'' of April 2023, the point at which its training data was collected. This means that neither model has been exposed to the \textit{Connections} puzzle, which debuted in July 2023, in its training data.

\section{Experiments}

\subsection{General Performance}
\label{sec:exp-1}

Our first experiment simply examines the capacity of our various models to solve the standard version of the \textit{Connections} puzzle. For each model, we allow up to 5 incorrect guesses and, in the case of the LLMs, up to 5 invalid guesses. Our baseline model is deterministic, so we evaluate it only once on each puzzle. For the LLMs, we set the sampling temperature to 0 and repeat each puzzle with three different random seeds. 
\footnote{We note that despite using greedy sampling and setting the random seed, the behavior of OpenAI langauge models is still not fully deterministic---more details are available at \href{https://platform.openai.com/docs/guides/text-generation/reproducible-outputs}{https://platform.openai.com/docs/guides/text-generation/reproducible-outputs}}
For each model, we compute its average ``success rate''---the proportion of puzzles and individual categories correctly solved. For the LLMs, we average over both puzzles and random seeds. We plot these success rates by model in \autoref{fig:avg_success_rates} (right three column groups).

To get a better sense of the behavior of the sentence embedding baselines, we perform an additional experiment on our best performing baseline model (\textsc{MPNet}) in which we allow up to 500 incorrect guesses for each puzzle. We then measure the proportion of puzzles and individual categories solved as a function of the allowed number of guesses. We present these results in \autoref{fig:baseline_extra}. Due to constraints on cost and computational budget, we were not able to perform this experiment with the LLMs.

\subsection{Effects of LLM Prompting}

\begin{figure*}
    \centering
    \includegraphics[width=2\columnwidth]{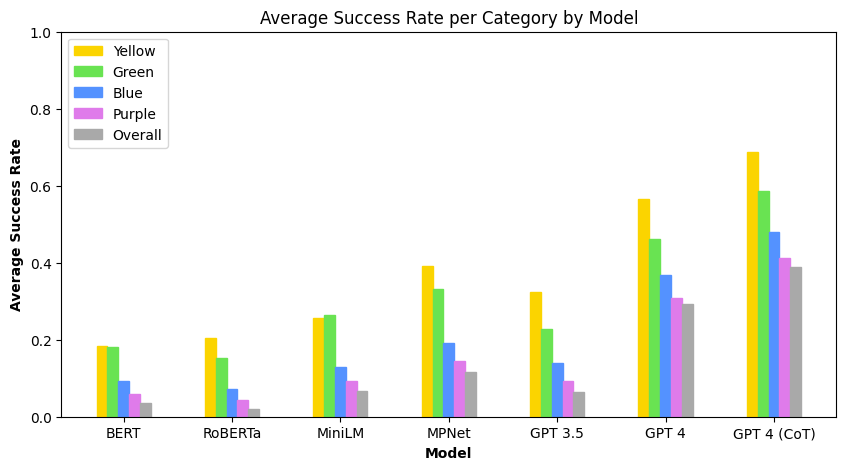}
    \caption{Average success rate across all puzzles and seeds for baseline models and LLMs, broken down by puzzle category (note that \textit{CoT} indicates the use of chain-of-thought prompting). \textbf{Categories increase in difficulty going from yellow to green to blue to purple.} First, we see that category difficulty generally aligns with success rate across models. Between models, we see that LLMs generally outperform the baseline, at best solving a sizeable proportion of puzzles (but not a majority). Finally, we see that chain-of-thought prompting provides a notable boost to LLM performance.}
    \label{fig:avg_success_rates}
\end{figure*}

For our second experiment, we turn our focus to the better-performing LLM (\textsc{gpt-4-turbo}) and examine the effects of prompting style on its performance. In particular, we augment the input to our model by adding \textit{chain-of-thought} prompting. Chain-of-thought is a prompting technique that instructs the model to work through a problem in multiple steps, explaining and justifying its reasoning at each stage. This approach has been shown to elicit plausible reasoning from language models and improve performance in a variety of domains~\cite{wei2022chain}. We prompt the model to first briefly explain the objectives of the puzzle, then propose a group and justify the inclusion of four words in that group, before returning its final answer. Aside from changes to the prompt, we use the same experimental setup as described in \autoref{sec:exp-1} and present the results in \autoref{fig:avg_success_rates} (right-most column group).

\subsection{Challenge Variant}

Our final experiment examines the performance of baseline and LLM solvers on the more challenging \textit{Connections} variant described in \autoref{sec:details}. We measure the performance of both our sentence-embedding baseline, \textsc{gpt-3.5-turbo}, and \textsc{gpt-4-turbo} using our default prompts, again keeping other experimental details unchanged (i.e. 5 incorrect and 5 invalid guesses allowed). 

\section{Results and Discussion}

\subsection{General Performance}

At the highest level, we find that the models are capable of solving some---but far from all---of the \textit{Connections} puzzles. When given an allowance of 5 incorrect guesses, the best-performing sentence embedding baseline (\textsc{MPNet}) has a $11.6\%$ success rate, \textsc{gpt-3.5-turbo} has a 6.43\% success rate, and \textsc{gpt-4-turbo} has a $29.2\%$ success rate (noting that LLM success rates are averaged over three seeds).
We apply Welch's $t$-test to compare \textsc{MPNet} to \textsc{gpt-3.5-turbo} and \textsc{gpt-3.5-turbo} to \textsc{gpt-4-turbo} and find both differences are significant ($p < 0.02$ and $p < 1.26 * 10^{-31}$, respectively).

When examining performance by category, we find that the ordering of category difficulty as specified by the puzzle corresponds exactly with the ordering of success rate across all models except \textsc{BERT} (i.e. the ``tricky'' purple category is solved less frequently than the ``simple'' yellow category). We perform a dependent $t$-test between each sequential category (i.e. between yellow and green, green and blue, ...) and find that the decrease in performance is significant across both LLMs for all colors ($p < 2.972 * 10^{-4}$) and significant for baselines except between the yellow and green categories (all models $p > 0.05$) and, for \textsc{BERT} and \textsc{RoBERTa}, between blue and purple categories.

Another interesting effect is that the performance of the LLMs depends greatly on whether its initial guess is correct. In \autoref{tab:first_guess} we present the averaged success rate for each LLM based on whether its initial guess was correct, incorrect, or \textit{nearly correct} (i.e. containing three out of the four words in a category). Across all models, performance decreases substantially if the initial guess is incorrect or nearly correct compared to general performance ($p < 7.14 * 10^{-4}$) except for \textsc{gpt-3.5-turbo} when the initial guess is nearly correct. This is at least in part because the models fall into ``rabbit holes'' where they expend all of their allowed guesses in attempting to solve a single nearly correct category.

Generally, we find that the LLMs are often stumped by categories which involve non-semantic properties of words (e.g. \textit{LETTER HOMOPHONES: are, why, sea, queue}), abstract features (e.g. \textit{MEMBERS OF A SEPTET: sea, sister, sin, wonder}), or usage in context (e.g. \textit{\_\_\_ PAPER: butcher, scrap, toilet, wax}). This raises an interesting question about the information represented by the learned latent vectors for each word or token. Standard training objectives seem sufficient to learn embeddings that reliably encode straightforward word properties but perhaps struggle to capture these more subtle groupings, potentially pointing towards the benefits of multi-modal models or explicit access to a database of world knowledge.

\begin{figure*}
    \centering
    \includegraphics[width=2\columnwidth]{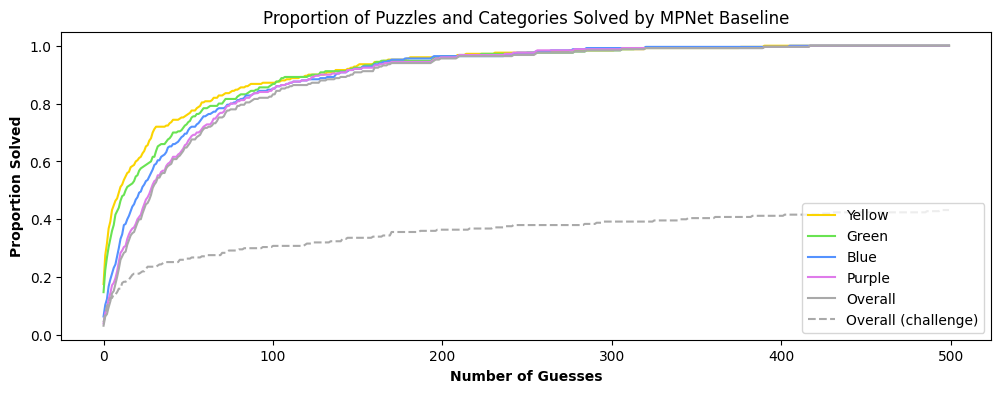}
    \caption{Proportion of all categories and overall puzzles solved by the \textsc{MPNet} sentence embedding baseline with an increasing amount of allowed guesses. Notably, half of all puzzles are solved within 29 incorrect guesses and all puzzles are solved within 417 guesses. On the challenge variant, only 108 puzzles are solved within 500 incorrect guesses.}
    \label{fig:baseline_extra}
\end{figure*}

\begin{table}[t]
    \centering
    \caption{Average success rates for LLMs on the standard Connections puzzle based on correctness of initial guess}
    \adjustbox{max width=\linewidth}{
        \begin{tabular}{lccc}
        \toprule
         & \multicolumn{3}{c}{\textbf{Success Rate when First Guess is...}} \\
        \textbf{Model} &  \textit{Correct} & \textit{Nearly Correct} & \textit{Incorrect}\\
        \midrule
        GPT-3.5 Turbo &  15.98\% & 4.79\% & 1.16\% \\
        GPT-4 Turbo &  50.56\% & 13.70\% & 2.40\% \\
        GPT-4 Turbo (CoT) &  58.11\% & 23.08\% & 14.41\% \\
        \bottomrule
        \end{tabular}
    }
    \label{tab:first_guess}
\end{table}

This hypothesis is somewhat belied by our additional baseline results, however. We find that the straightforward sentence embedding approach solves every one of the 150 puzzles within 417 of the allowed 500 incorrect guesses, and it solves over half within 29 guesses. There are over 2.5 million possible four-group guesses for a given \textit{Connections} puzzle (dropping to 5,775 once the first group has been found), indicating that this is not a result of brute-forcing the puzzle. So, it seems likely that the sentence embeddings capture, at least weakly, the semantic links that tie \textit{Connections} groups together. However, a substantial gap still remains between the apparent amount of search required in embedding space (i.e. number of incorrect guesses) and the amount needed by LLM or human players.

\subsection{Effects of LLM Prompting}

Our results here are relatively straightforward: chain-of-thought prompting improves the performance of the \textsc{gpt-4-turbo} model both overall (from $29.2\%$ averaged success rate to $38.93\%$ averaged success rate, $p < 6.77*10^{-5}$) and within each difficulty category ($p < 2.66*10^{-5}$), aligning with previous findings in other domains \cite{wei2022chain}. One interesting aspect of chain-of-thought prompting is that it causes the model to provide a name and a justification for each category it guesses. This opens the door for a further analysis---do the LLMs generate the ``right'' categories when they correctly solve a puzzle, and do their justifications make sense? We leave these interesting questions for future work. 

\subsection{Challenge Variant}
\label{sec:challenge-results}
We report our results in \autoref{tab:challenge}. For the \textsc{MPNet} baseline, the performance does not change significantly between the default puzzle and challenge variant ($p > 0.68$) with the standard allowance of incorrect guesses. However, we see in \autoref{fig:baseline_extra} that a large decrease in performance appears with a larger allowance of incorrect guesses. 
The story for the LLMs is mixed -- performance for \textsc{gpt-3.5-turbo} drops on the challenge variant and is significant ($p < 4.36 * 10^{-6}$), while for \textsc{gpt-4-turbo} the apparent drop in performance is \textit{not} significant ($p > 0.11$). It is possible that, like with the \textsc{MPNet} baseline, a larger number of allowed guesses would reveal a larger difference in performance. Chain of thought prompting appears less helpful in this regime, with the performance of \textsc{gpt-4-turbo} dropping significantly to $23.46\%$ ($p < 7.83 * 10^{-11}$). This indicates that while the challenge variant could act as a more difficult benchmark task in the event that models quickly become able to completely solve the standard puzzle, its impact on performance is somewhat muddled.

We note that our success rates are much lower than the $65\%$ reported by a prior investigation of the first 60 \textit{Connections} puzzles, in an experiment that also uses an older version of the GPT-3.5 model and allows only a single guess for each puzzle \cite{perowne2023connections}. We investigate this discrepancy by replicating their experiment, using their prompt and the \textsc{gpt-3.5-turbo} model. On the first 60 puzzles, our replication fares substantially worse, achieving only a $3.33\%$ success rate. One potential account of this difference lies in the way the puzzle is presented to the model. Perowne and Iancu provide the sixteen words of each puzzle in grouped order (i.e. all words belonging to the yellow category, followed by all words belong to the green category, followed by...), whereas our interface randomizes the ordering of the words for each puzzle. When we similarly sort the words, our replication achieves an identical 65\% success rate, indicating that word ordering was likely the driving factor of the model's deceptively high performance.

\begin{table}[t]
    \centering
    \caption{Average success rates for baseline model and LLMs on the standard Connections puzzle and challenge variant}
    
    \begin{tabular}{lcc}
    \toprule
    \textbf{Model} &  \textbf{Success (Base)} & \textbf{Success (Challenge)} \\
    \midrule
    MPNet &  11.6\% & 12.8\% \\
    GPT-3.5 Turbo &  6.42\% & 1.73\% \\
    GPT-4 Turbo &  29.2\% & 25.56\% \\
    GPT-4 Turbo (CoT) &  38.93\% & 23.46\% \\
    \bottomrule
    \end{tabular}
    \label{tab:challenge}
\end{table}

\section{Future Work}

While the LLMs examined here are capable of solving many puzzles (and in the case of GPT-4 exhibit relatively high success rates overall),  there are a number of approaches that could offer further improvement. For instance, a dedicated training set of puzzles could be leveraged to either directly learn a solver or to iteratively refine a set of prompts (an approach that seems particularly promising given the substantial performance increase afforded by the relatively simple inclusion of chain-of-thought prompting). In addition, the \textit{Connections} puzzle could be a useful way to measure the effectiveness of integrating explicit knowledge bases with LLMs, as databases like \textsc{WordNet} \cite{fellbaum2010wordnet} seem likely to contain a wealth of puzzle-relevant information that may be more difficult to access through trained word embeddings.

In another direction, we are interested in the potential of using LLMs to \textit{generate} novel linguistic puzzles. This task can be seen as strictly harder than solving such a puzzle, since it requires creativity and a sense of design in addition to the semantic understanding required for acting as a solver. The \textit{Connections} puzzle, being both straightforward in its rules and wide-ranging in its difficulty, presents an appealing test-bed for such generative approaches which could be then be applied to assist designers in other games or domains.

Finally, there are many interesting questions to examine in comparing the ways that human players and LLMs solve \textit{Connections} puzzles. We know that category difficulty aligns with LLM success rate, but do LLMs solve puzzles in the same order as humans? Are distractors that are challenging for LLMs also challenging for human players, and vice-versa? More generally, studying the ways in which LLM and human player behaviors differ could shed light on the differences in their underlying representations of words and meanings.

\section{Conclusion}

We study the extent to which modern NLP systems trained on vast amounts of linguistic data contain semantic knowledge and abstract associations by leveraging them to solve \textit{Connections}, a recent and widely popular word puzzle game. We find that large language models in the GPT family are able to solve these puzzles with moderate reliability, indicating that the task is possible but remains a formidable challenge. We also replicate recent findings showing that prompting LLMs to reason about their outputs improves their performance.

The unique nature of the \textit{Connections} puzzle---which involves identifying often uncommon or lateral associations among groups of words---and its large player base are evidence of its potential as a benchmark for a distinct and interesting kind of linguistic reasoning. We expect that the \textit{Connections} puzzle (and the more challenging variant studied here) can serve as a helpful tool among other games and tasks for studying future language models and linguistic systems.

\bibliographystyle{IEEEtran}
\bibliography{bibliography}

\appendix

\subsection*{Ethical Statement}

Applications of language models and word embeddings are capable of reproducing and reinforcing biases present in their training corpuses. While the scope of our approach is limited to \textit{Connections} puzzles, any investigation that aims to study these models should acknowledge their potential biases. Particular care should be paid to any future work that directly interfaces a language model with players or uses language model output as part of a puzzle-generating process.

\subsection*{LLM Prompts}
\label{sec:prompts}

\lstdefinelanguage{puzzle_prompts}{
  morekeywords={PUZZLE_WORDS, CHAIN_OF_THOUGHT_PROMPT, CATEGORY_NAME, CATEGORY_COLOR}, 
  keywordstyle=\color{blue}\bfseries, 
  sensitive=false, 
  breaklines=true, 
  columns=fullflexible,
  basewidth = {.5em},
  breakindent = {0em},
  tabsize=2,
  aboveskip=1em,
  belowskip=1em,
  comment=[l]{//}, 
  commentstyle=\color{gray}\ttfamily, 
  basicstyle=\footnotesize\ttfamily, 
  showstringspaces=false, 
  morestring=[b]", 
  morestring=[b]' 
}

\definecolor{mygray}{rgb}{0.5,0.5,0.5}
\definecolor{lightgray}{rgb}{0.9,0.9,0.9}

\lstset{
  backgroundcolor=\color{lightgray},
  basicstyle=\ttfamily\small,
  breaklines=true,
  frame=none, 
  language=puzzle_prompts, 
}

\begin{center}
  \doublebox{ 
    \begin{minipage}{0.9\columnwidth}
      \textbf{Initial Prompt for Iterative Solving}
      \begin{lstlisting}

I want you to solve a daily word puzzle that finds commonalities between words. There are 16 words, which form 4 groups of 4 words. Each group has some common theme that links the words. You must use each of the 16 words, and use each word only once.
    
Each group of 4 words are linked together in some way. The connection between words can be simple. An example of a simple connection would be "types of fish": Bass, Flounder, Salmon, Trout. Categories can also be more complex, and require abstract or lateral thinking.
An example of this type of connection would be "things that start with FIRE": Ant, Drill, Island, Opal...
      \end{lstlisting}
    \end{minipage}
  }
\end{center}

\begin{center}
  \doublebox{ 
    \begin{minipage}{0.9\columnwidth}
      \textbf{Initial Prompt for Iterative Solving (Continued)}
      \begin{lstlisting}
...Provide the one group you are most sure of as your final answer. I will enter this into the puzzle and give you feedback: I will tell you whether it is correct, incorrect, or nearly correct (3/4 words).
Then we will continue until the puzzled is solved, or you lose.

Format your answer as:
GROUP NAME: [WORD, WORD, WORD, WORD]

Some rules:
{CHAIN_OF_THOUGHT_PROMPT}- Give your final answer in the format described above. Do not add any additional text to your final answer, just the group name and the 4 words. 

Here are the starting 16 words:
{PUZZLE_WORDS}
      \end{lstlisting}
    \end{minipage}
  }
\end{center}


\begin{center}
  \doublebox{ 
    \begin{minipage}{0.9\columnwidth}
      \textbf{Chain-of-thought prompt for Iterative Solving}
      \begin{lstlisting}
- First, briefly summarize the rules and objective of the puzzle (in no more than 50 words)
- Next, come up with a category to which four of the words belong and briefly explain why you think they belong to that category:
      \end{lstlisting}
    \end{minipage}
  }
\end{center}

\begin{center}
  \doublebox{ 
    \begin{minipage}{0.9\columnwidth}
      \textbf{Correct Guess Prompt for Iterative Solving}
      \begin{lstlisting}
The response from the game was: Correct! The category was {CATEGORY_NAME}. Diffulty: {CATEGORY_COLOR}

Continue to solve the puzzle. 
Format your answer as:
GROUP NAME: [WORD, WORD, WORD, WORD]

Here are the remaining words:
{PUZZLE_WORDS}
      \end{lstlisting}
    \end{minipage}
  }
\end{center}

\begin{center}
  \doublebox{ 
    \begin{minipage}{0.9\columnwidth}
      \textbf{Nearly Correct Guess Prompt for Iterative Solving}
      \begin{lstlisting}
The response from the game was: Nearly Correct. Three of your words are in a group, but one is not in the same group.

Continue to solve the puzzle. Again, provide one group you are most certain of. MAKE SURE YOU DON'T REPEAT ANY OF YOUR PREVIOUS GUESSES.
Format your answer as:
GROUP NAME: [WORD, WORD, WORD, WORD]

Here are the remaining words:
{PUZZLE_WORDS}
      \end{lstlisting}
    \end{minipage}
  }
\end{center}

\begin{center}
  \doublebox{ 
    \begin{minipage}{0.9\columnwidth}
      \textbf{Incorrect Guess Prompt for Iterative Solving}
      \begin{lstlisting}
The response from the game was: Incorrect guess.

Let's continue to solve the puzzle. MAKE SURE YOU DON'T REPEAT ANY OF YOUR PREVIOUS GUESSES.
Format your answer as:
GROUP NAME: [WORD, WORD, WORD, WORD]

Here are the remaining words:
{PUZZLE_WORDS}
      \end{lstlisting}
    \end{minipage}
  }
\end{center}

\begin{center}
  \doublebox{ 
    \begin{minipage}{0.9\columnwidth}
      \textbf{Invalid Guess Prompt for Iterative Solving}
      \begin{lstlisting}
The response from the game was: Invalid guess. Please try again.

Your answer wasn't formatted correctly. Try again, and follow the formatting instructions carefully.
Format your answer as:
GROUP NAME: [WORD, WORD, WORD, WORD]

Here are the remaining words:
{PUZZLE_WORDS}
      \end{lstlisting}
    \end{minipage}
  }
\end{center}


\begin{center}
  \doublebox{ 
    \begin{minipage}{0.9\columnwidth}
      \textbf{Initial Prompt for All-in-One Solving}
      \begin{lstlisting}
I want you to solve a daily word puzzle that finds commonalities between words. There are 16 words, which form 4 groups of 4 words. Each group has some common theme that links the words. You must use each of the 16 words, and use each word only once. Each group of 4 words are linked together in some way. The connection between words can be simple. An example of a simple connection would be "types of fish": Bass, Flounder, Salmon, Trout. Categories can also be more complex, and require abstract or lateral thinking.
An example of this type of connection would be "things that start with FIRE": Ant, Drill, Island, Opal.

Format your final answers as:
GROUP 1 NAME: WORD, WORD, WORD, WORD
GROUP 2 NAME: WORD, WORD, WORD, WORD
GROUP 3 NAME: WORD, WORD, WORD, WORD
GROUP 4 NAME: WORD, WORD, WORD, WORD

Replace each GROUP NAME with a name for the group you create.

Some rules:
- Give your final answers in the format described above. Put each group on a separate line. Do not add any additional text to your final answer, just the group name and the 4 words. 

Here are the starting 16 words:
{PUZZLE_WORDS}
      \end{lstlisting}
    \end{minipage}
  }
\end{center}


\begin{center}
  \doublebox{ 
    \begin{minipage}{0.9\columnwidth}
      \textbf{Incorrect Guess Prompt for All-in-One Solving}
      \begin{lstlisting}

The response from the game was: Incorrect guess.

Let's continue to solve the puzzle. MAKE SURE YOU DON'T REPEAT ANY OF YOUR PREVIOUS GUESSES.

Format your final answers as:
GROUP 1 NAME: WORD, WORD, WORD, WORD
GROUP 2 NAME: WORD, WORD, WORD, WORD
GROUP 3 NAME: WORD, WORD, WORD, WORD
GROUP 4 NAME: WORD, WORD, WORD, WORD

The remaining words are:
{PUZZLE_WORDS}
      \end{lstlisting}
    \end{minipage}
  }
\end{center}

\begin{center}
  \doublebox{ 
    \begin{minipage}{0.9\columnwidth}
      \textbf{Invalid Guess Prompt for All-in-One Solving}
      \begin{lstlisting}
The response from the game was: Invalid guess. Please try again.

Your answer wasn't formatted correctly. Try again, and follow the formatting instructions carefully.

Format your final answers as:
GROUP 1 NAME: WORD, WORD, WORD, WORD
GROUP 2 NAME: WORD, WORD, WORD, WORD
GROUP 3 NAME: WORD, WORD, WORD, WORD
GROUP 4 NAME: WORD, WORD, WORD, WORD

The remaining words are:
{PUZZLE_WORDS}
      \end{lstlisting}
    \end{minipage}
  }
\end{center}

\begin{center}
  \doublebox{ 
    \begin{minipage}{0.9\columnwidth}
      \textbf{Modified ChatGPT Prompt used by \cite{perowne2023connections} (formatting instructions added)}
      \begin{lstlisting}
Find 4 groups, each of 4 words that share something in common, out of 16 words. I want to use them to solve a daily word puzzle that finds commonalities between words. The game is a new puzzle featured in The New York Times, inspired by crosswords. You have to use all those 16 words I give you and each word only once.
Format your final answers as:
GROUP 1 NAME: [WORD, WORD, WORD, WORD]
GROUP 2 NAME: [WORD, WORD, WORD, WORD]
GROUP 3 NAME: [WORD, WORD, WORD, WORD]
GROUP 4 NAME: [WORD, WORD, WORD, WORD]

Below are my 16 words: 
{PUZZLE_WORDS}
      \end{lstlisting}
    \end{minipage}
  }
\end{center}

\end{document}